\documentclass[letterpaper, 10 pt, conference]{ieeeconf}  % Comment this line out if you need 

\IEEEoverridecommandlockouts                              % This command is only needed if 
\usepackage{url}
\usepackage{cite}
\usepackage{times}
\usepackage{graphicx}
\usepackage{amsmath}
\usepackage{amssymb}
\usepackage{enumerate}
\usepackage{mathtools}
\usepackage{amsmath}
\usepackage{algorithm}  
\usepackage{algorithmic}

\usepackage{subfigure}
\usepackage{float}
\usepackage{textcomp}
\usepackage{bbding}
\usepackage{pifont}
\usepackage{leftidx}
\usepackage{bm}
\newcommand\degree{^\circ}

\title{\LARGE \bf
FlowNorm: A Learning-based Method for Increasing Convergence Range of Direct Alignment
}

\author{Ke Wang, Kaixuan Wang, and Shaojie Shen % <-this % stops a space
\thanks{The authers are with Department of Electronic and Computer Engineering , 
        Hong Kong University of Science and Technology,
        {\tt\small kwangbd@connect.ust.hk}}%
}

\begin{document}

\maketitle
\thispagestyle{empty}
\pagestyle{empty}

%%%%%%%%% ABSTRACT
\begin{abstract}

Many approaches have been proposed to estimate camera poses by directly minimizing photometric error. However, due to the non-convex property of direct alignment, proper initialization is still required for these methods. Many robust norms (e.g. Huber norm) have been proposed to deal with the outlier terms caused by incorrect initializations. These robust norms are solely defined on the magnitude of each error term. In this paper, we propose a novel robust norm, named FlowNorm, that exploits the information from both the local error term and the global image registration information. While the local information is defined on patch alignments, the global information is estimated using a learning-based network. Using both the local and global information, we achieve an unprecedented convergence range in which images can be aligned given large view angle changes or small overlaps. We further demonstrate the usability of the proposed robust norm by integrating it into the direct methods DSO and BA-Net, and generate more robust and accurate results in real-time.

\end{abstract}

\section{Introduction}

Direct methods are widely used to solve visual odometry and monocular stereo problems~\cite{RefWorks:doc:5c72ad8ee4b04f2c904d8176,RefWorks:doc:5c7379a0e4b02dd985839ef3,RefWorks:doc:5c737910e4b085fcc7aa11d0,RefWorks:doc:5c72ae23e4b0819eb0af23dc}. By directly minimizing the photometric error between pixels in the source frame and the target frame, camera poses and scene geometry can be estimated in the joint optimization process. Compared with indirect methods~\cite{RefWorks:doc:5c7377a5e4b063c5906211b8,RefWorks:doc:5c7376d7e4b0b6d3b1ea9dc7,RefWorks:doc:5c73787ce4b053a1d93364df,RefWorks:doc:5c7377dfe4b04f2c904db140}, which solve the problem by minimizing the reprojection error between matched sparse features, direct methods avoid the pre-processed feature matching step and can utilize more pixels in the image. However, intensity-based optimization is prone to local minima due to the non-convex property of complex images.

Recent years, many approaches have been proposed to expand the convergence range of direct methods. SVO~\cite{RefWorks:doc:5d7df305e4b064b49fd56169} combines matched feature points with photometric optimization. However, although matched features can provide pose initialization for further optimization, they rely on textures of the environment and are prone to outliers. With the help of learning-based methods, many researchers have proposed networks~\cite{RefWorks:doc:5c73d9a4e4b04f2c904dc779},~\cite{Stumberg2019GNNetTG} to generate smooth feature maps for direct optimization. Compared with the image intensity domain, optimization on feature maps shows advantages in convergence ranges. For example, BA-Net~\cite{RefWorks:doc:5c73d9a4e4b04f2c904dc779} can estimate camera poses given images with small overlaps. LS-Net~\cite{RefWorks:doc:5c73d59ae4b053a1d933774a} uses an end-to-end trained network as a solver for two-frame monocular stereo problems. Learning-based methods achieve superior performance on evaluation datasets, such as RGB-D datasets or the KITTI dataset, but have not been widely used on robotic platforms. The reason may be the limited computation resources of general robotic platforms and the diversity of robotic application scenes.

\begin{figure}[t]
  \begin{center}
  \includegraphics[width=1.0\linewidth]{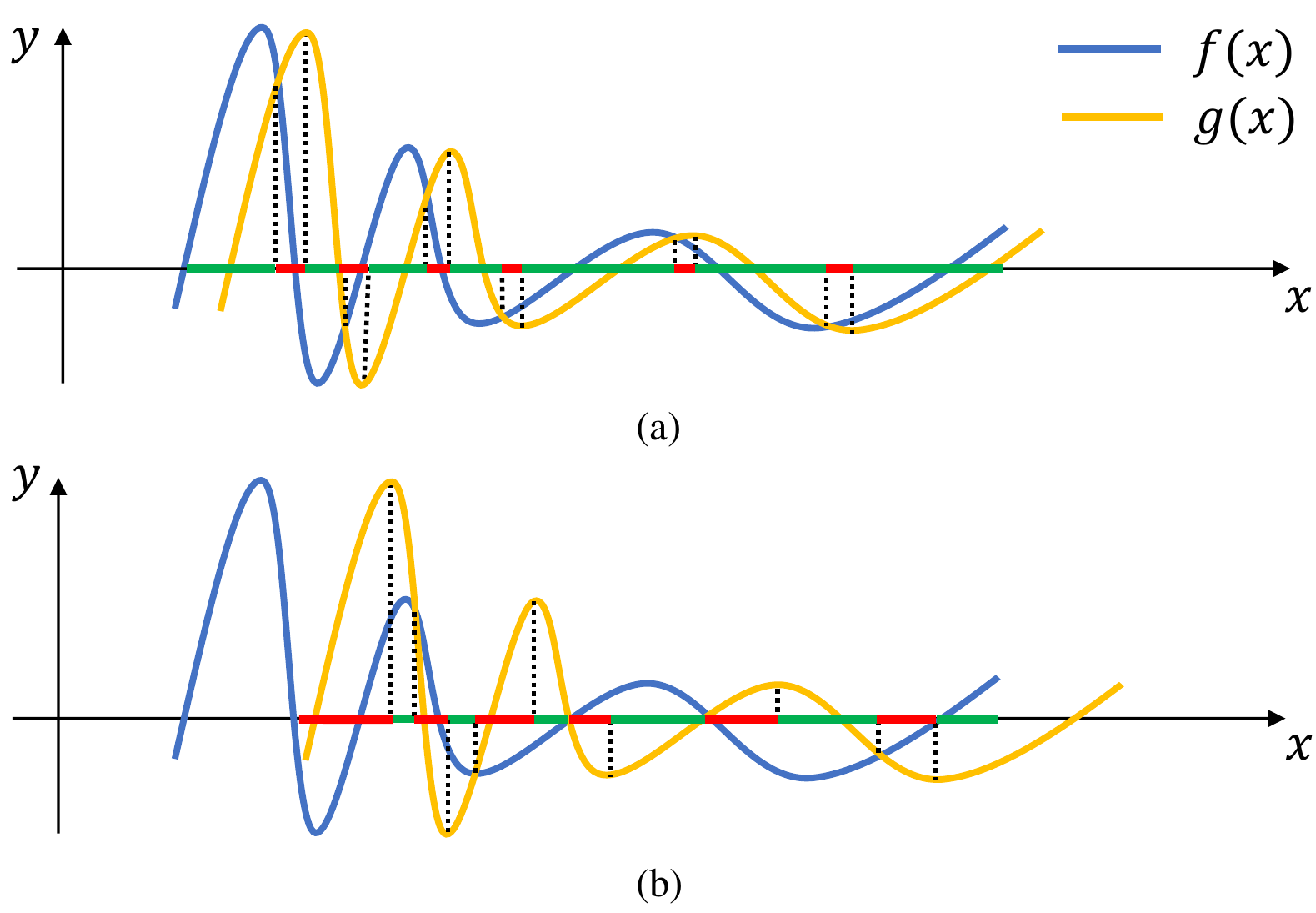}
  \end{center}
    \caption{A simple example to show the different contributions of points in aligning two functions: $\text{argmin}_t \sum_x \lVert g(x+t)-f(x) \rVert_2$. Point $x$ with $(g(x+t)-f(x))g'(x+t) < 0$ contributes to the optimization of $t$ and is marked in green, while $x$ with $(g(x+t)-f(x))g'(x+t) > 0$ counteracts the optimization and is marked in red. As shown in (a), with good initialization, most of the points are positive to the optimization. However, with worse initialization, negative points make the optimization fall into local minima.}

  \label{fig:CorrespondenceCurve}
\end{figure}

One of the contributions of the paper is a study of the direct optimization process followed by the design of a robust norm for the optimization. Due to the problem of non-convexity, during the photometric minimization (or feature consistency minimization in learning-based methods), not all pixels contribute to the convergence. The difference in pixels depends on both the local texture and the global pose initialization which establishes pixels correspondences. We illustrate the convergence problem in Fig.~\ref{fig:CorrespondenceCurve}. As shown, for good initializations, most of the correspondences contribute to the final estimation. However, given a bad initialization, most of the correspondences will suppress the convergence. Poor correspondences make the optimization fall into local minima. Based on this observation, we propose the flow norm that combines a low-accuracy optical flow prediction network to distinguish which correspondences will suppress the convergence of direct alignment. 

%%%%%%%%% BODY TEXT

%One of the contributions in the paper is a study of the optimization process. Due to the highly non-convex problem, during the photometric minimization (or feature consistency minimization in learn-based methods), not all pixels contributes to the convergence. The difference of pixels depends on both the local texture and the global pose initialization which establishes pixels correspondences. We illustrate the converge problem in Fig.~\ref{fig:CorrespondenceCurve}. As shown, for good initializations, most of the correspondences contribute to the final estimation. However, given bad initialization, most of the correspondences will suppress the convergence, thus poor correspondences make the optimization fall into local minimums. Based on the observation, we propose the flow norm that combines an optical flow prediction network to distinguish which correspondences will suppress the convergence of direct alignment. 

In summary, the contributions of our paper are the following:
\begin{itemize}
\item{
We propose a new norm to expand the convergence range of the traditional nonlinear solver for the direct alignment problem.}
\item{To the best of our knowledge, the proposed method is the first that can distinguish which correspondence will suppress solver convergence in the direct alignment situation.}
\item{We build FlowNorm versions of DSO and BA-Net, and the FlowNorm DSO retains the real-time property.}
\end{itemize}
To demonstrate the effectiveness of our method, we evaluate it on the SceneNN dataset~\cite{RefWorks:doc:5c7bd892e4b05f3521397a13},
TUM-MonoVO dataset~\cite{RefWorks:doc:5d7dfb06e4b054c6b854fc6a} and ICL-NUIM dataset~\cite{RefWorks:doc:5d7dfbd6e4b064b49fd561f2}, showing that FlowNorm versions  consistently outperform the original versions.

\section{Related Work}
%
%\noindent {\bf Traditional methods for direct alignment} SVO~\cite{RefWorks:doc:5d7df305e4b064b49fd56169} is a pioneering work that tracks a monocular camera in real-time using direct alignment algorithm, which takes an initialization from the feature matched method as the initial pose of the successive direct alignment. LSD-SLAM~\cite{RefWorks:doc:5c737910e4b085fcc7aa11d0} modified the bundle adjustment of backend from $se(3)$ to $sim(3)$, thereby explicitly detecting scale-drift. DSO~\cite{RefWorks:doc:5d7df900e4b07e3e85c82d02} is the baseline work of direct alignment, which jointly optimizes all model parameters, including geometry represented as inverse depth in a reference frame and
%camera motion. What's more, DSO integrates a full photometric calibration, accounting for exposure time, lens vignetting, and non-linear response functions. 
%Although these methods feature real-time efficiency and high accuracy, they rely on incrementally tracking the camera poses to ensure large overlaps and proper initialization.

Semi-dense visual odometry~\cite{RefWorks:doc:5c72ae23e4b0819eb0af23dc} is a pioneering work that tracks a monocular camera in real-time using direct alignment algorithm. SVO~\cite{RefWorks:doc:5d7df305e4b064b49fd56169} uses matched features to calculate an initialization pose for joint optimization. Following the idea of the direct method, Engel proposed LSD-SLAM~\cite{RefWorks:doc:5c737910e4b085fcc7aa11d0} ,which solves the camera pose using keyframes with depth values. DSO~\cite{RefWorks:doc:5d7df900e4b07e3e85c82d02} is the baseline direct alignment work, which jointly optimizes all model parameters, including geometry represented as inverse depth and camera motion. DSO further integrates a full photometric calibration, accounting for exposure time, lens vignetting, and non-linear response functions. Although all these methods feature real-time efficiency and high accuracy, they rely on incrementally tracking the camera poses to ensure large overlaps and proper initialization.

To increase the convergence range of the direct methods, many learning-based methods have been proposed to replace the intensity map with feature maps. BA-Net \cite{RefWorks:doc:5c73d9a4e4b04f2c904dc779} formulates Bundle Adjustment (BA) as a differentiable layer and utilizes a standard encoder-decoder network to generate the feature map and depth map. Camera poses and depth maps are optimized by minimizing the feature consistency between projected pixels. Benefiting from the generated feature maps, BA-Net expands the convergence range of direct alignment. GN-Net~\cite{Stumberg2019GNNetTG} uses a novel Gauss-Newton loss for training deep feature maps. The direct alignment in GN-Net, based on minimizing the feature metric error, achieves robust performance under dynamic lighting or weather changes. These two approaches nicely combine traditional direct alignment and deep learning techniques. 
%(Our method does not conflict with the feature metric methods and can further expand the convergence range of these methods. 

LS-Net \cite{RefWorks:doc:5c73d59ae4b053a1d933774a} uses an end-to-end trained network to replace the traditional nonlinear solver. Given a photometric error map and a Jacobian matrix, LS-Net estimates the updated depth map and camera motion. Although it achieves impressive results on datasets, the generalization ability of LS-Net has not been demonstrated.

In this paper, we propose a different solution that improves the robustness of direct optimization. The core of the contribution is a robust norm that distinguishes error terms using both local and global information. Different from most of the learning-based methods that use a heavy network to generate high-dimensional feature maps, we utilize a light-weight network to improve both the robustness and accuracy of the state-of-the-art methods, with an overhead of only $14\ ms$.

\section{Direct Alignment Revisited}
\label{Revisited}
Before introducing our enhanced direct alignment algorithm, we revisit the classic direct alignment to give a better understanding of where difficulties lie, and why our method is desirable. We only introduce the most relevant content, and refer the readers to \cite{RefWorks:doc:5d7df900e4b07e3e85c82d02} for a more comprehensive introduction.

%\noindent {\bf Notation.} $I_s$ and $I_t$ is a target/source image pair, For a point $p_i$ in source image $I_s$, $p_i^{'}$ and $\dot{p}^{'}_i$ are the real corresponding postion of $p_0$ and estimating position by current parametors $\mathcal{X}$ in the target image $I_t$ respectively. We take one pixel $p_i$ ,  The predicted flow from $I_s$ to $I_t$ is $f$ and $f_i$ denote the flow of $p_i$, thus, $\widehat{p}^{'}_i=f_i+p_i$ is the predicted corresponding postion of $p_i$ in $I_t$. What's more, we take $\overrightarrow{dp_i^{'}}$ to denote the derivative of residual $e_i$ with respect to $p^{'}_i$, which is define as 

Given a target/source image pair $I_s$ and $I_t$, the direct alignment problem is formulated as estimating the relative transformation $\bm{T}$ between the image pair, and $d_i\in D=\{d_i|i=1\cdots N\}$, which are the depths of the pixels $\bm{p_{si}}\in P=\{\bm{p_{si}}|i=1\cdots N\}$ at the image $I_s$. Let $\bm{\mathcal{X}}=\left\{\bm{T},\bm{D}\right\}$ and we can estimate $\bm{\mathcal{X}}$ by minimizing the norm of the photometric error
\begin{equation}
\begin{aligned}
\hat{\bm{\mathcal{X}}}=\mathop{\arg\min}_{\bm{\mathcal{X}}}\sum_{i=1}^{N}\left|e_{i}(\bm{\mathcal{X}})\right|,
\end{aligned}
\label{f1}
\end{equation}
where $\left|\cdot\right|$ donates the L1 norm or Huber norm of a vector, $N$ is the number of selected pixels, and the photometric error 
\begin{equation}
\begin{aligned}
e_{i}(\bm{\mathcal{X}})&=I_t(\bm{p_{ti}^{'}})-I_s(\bm{p_{si}})
\end{aligned}
\label{f100}	
\end{equation}
measures the intensity difference between the $ith$ pixel $\bm{p_{si}}$ at $I_s$ and its corresponding 
pixel $\bm{p_{ti}^{'}}$ at $I_t$. $\bm{p_{ti}^{'}}$ is computed by the projection function
%\pi(p_i,T,d_i) 
\begin{equation}
\begin{aligned}
\bm{p_{ti}^{'}}=\pi(\bm{p_{si}},\bm{T},d_i) 
=s{\bm{K}\bm{T}d_i\bm{K^{-1}}\bm{p_{si}}},
\end{aligned}
\label{f1001}	
\end{equation}
which projects 2D point $\bm{p_{si}}$ from $I_s$ to $I_t$, where $d_i$ is the depth value of $\bm{p_{si}}$ at $I_s$, $\bm{K}$ and $s$ are the camera's intrinsic matrix and a scale factor respectively. 

The general strategy to minimize Eq.~\eqref{f1} is the Gaussian-Newton (GN) or Levenberg-Marquardt (LM) algorithms\cite{RefWorks:doc:5c737a5ae4b063c5906211dc}. 
The GN and LM methods are both iterative methods. At the $jth$ iteration, the GN algorithm solves for an optimal update
\begin{equation}
\begin{aligned}
\Delta\bm{\mathcal{X}_j}=\bm{-(J_j{^T}J_j)^{-1}J_j{^T}E_j}.
\end{aligned}
\label{f100001}	
\end{equation}
%+\lambda diag(J^TJ)  and $\lambda$ is the damping factor
Here $\bm{E_j}=\left[e_{1}(\bm{\mathcal{X}_j}),e_{2}(\bm{\mathcal{X}_j}),\cdots,e_{N}(\bm{\mathcal{X}_j})\right]$, where $\bm{\mathcal{X}_j}$ is the initial parameters at the $jth$ iteration. Let $\bm{\delta}$ denotes a small $\bm{se(3)}$ pertubation around $\bm{\mathcal{X}_j}$, $\bm{J_j}$ is the Jacobian matrix of $\bm{E_j}$ with respect to $\bm{\delta}$. Let $\bm{p_{ti}^{'}}$ represent the projection position of $\bm{p_{si}}$ at $I_t$ based on the parameters $\bm{\mathcal{X}_j}$. The $ith$ row of $\bm{J_j}$ is
\begin{equation}
\begin{aligned}
\bm{J_j(i)} &= \left[\frac{\partial{e_i(\bm{\mathcal{X}_j})}}{\partial{I_t(\bm{p_{ti}^{'}})}} \frac{\partial{I_t(\bm{p_{ti}^{'}})}}{\partial{\bm{p_{ti}^{'}}}} \frac{\partial{\bm{p_{ti}^{'}}}}{\partial{\bm{\delta}}}\right],                             
\end{aligned}
\label{f5}
\end{equation}
where $\frac{\partial{e_i(\bm{\mathcal{X}_j})}}{\partial{I_t(\bm{p^{'}_{ti})}}}$ and $\frac{\partial{\bm{p^{'}_{ti}}}}{\partial{\bm{\delta}}}$ are smooth compared with the increment $\Delta\bm{\mathcal{X}_j}$. In contrast, $\frac{\partial{I_t(\bm{p^{'}_{ti}})}}{\partial{\bm{p^{'}_{ti}}}}$ is much less smooth. As found in DSO, $\frac{\partial{I_t(\bm{p^{'}_{ti}})}}{\partial{\bm{p^{'}_{ti}}}}$ is only valid in a 1-2 pixel radius. Hence the effective optimization requires that all parameters involved in computing $\bm{p^{'}_{ti}}$ should be initialized sufficiently accurately to be off by no more than 1-2 pixels. However, giving accurate initialization is difficulty when there is a large view change between $I_s$ and $I_t$.

\begin{figure}[t]
  \begin{center}
  \includegraphics[width=1.0\linewidth]{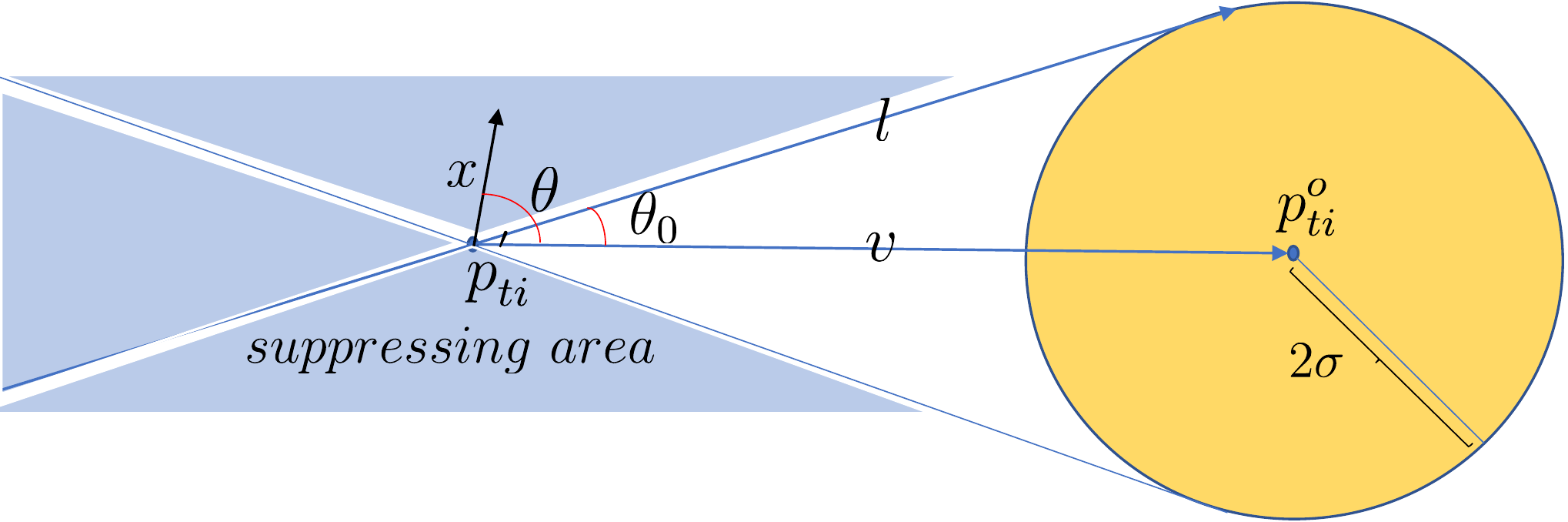}
  \end{center}
\vspace{-0.3cm}
    \caption{Illustration of $\theta$ and $\theta_0$, which are used in the definition of the flow norm. $\bm{x}$ is the derivative of the residual with respect to $\bm{p_{ti}^{'}}$, which totally depends on the local information. Conversely, $\theta_0$ relies on global information $\bm{p_{ti}^{'}}$, $\bm{p_{ti}^{o}}$ and $\sigma$.}
  \label{fig:ThetaAndTheta0}
\end{figure}

\section{Approach}
%(the difference between the projection position and the flow position)
To deal with the local minima problem of direct alignment, we design a flow norm to guide the non-linear solver to jump out from the local minimum. Assume we have a coarse optical flow map between the image pair. The key idea of the FlowNorm is to balance the local information (residual decreasing direction) and global optical flow information. Because the optical flow is coarse and unreliable, we just down-weight those correspondences whose residual decreasing directions are highly inconsistent with the corresponding flow positions. 
%The strategy avoids the noisy flow makes convergence situation worse and worse.

\subsection{Flow Norm}

Following the definition in Sect.~\ref{Revisited}, $\bm{F}$ denotes the computed coarse flow map between $I_s$ and $I_t$, $\bm{P_t^o}$ represents the flow positions computed by $\bm{P_t^o}=\bm{P_s}+\bm{F}$, and $\bm{P^{'}_t}$ is the projection position of $\bm{P_s}$ at $I_t$ based on the current relative pose $\bm{T}$. $\bm{p_{si}}$, $\bm{p_{ti}^o}$ and $\bm{p_{ti}^{'}}$ denote the $ith$ item of $\bm{P_s}$,$\bm{P_t^{o}}$ and $\bm{P^{'}_t}$ respectively. The flow norm of residual $e_i$ is defined as
\begin{equation}
L(\bm{p_{si}},\bm{p_{ti}^{o}},\bm{p^{'}_{ti}}, e_i) = \\
\left\{
             \begin{array}{lr}
             e_i,  \ \ \ \ \ \ \ \ \  \left|\bm{p_{ti}^o} - \bm{p_{ti}^{'}}\right|_2 \le 2\sigma \\
             e_i, \ \ \ \ \ \ \ \ \ \ \cos{\theta} \le \cos{\theta_0}\ \\ 
             (\frac{\cos{\theta}+1}{\cos{\theta_0}+1})e_i,  \ \ \ \cos\theta_0 < \cos{\theta},
             \end{array}
\right.
\end{equation}
where $\sigma$ is the variance of the computed flow (the method for computing $\sigma$ is described in Sect.~\ref{Shrinked}) and $\left|\cdot\right|_2$ denotes the L2 norm of the vector. $\bm{v} = \bm{p_{ti}^{'}} - \bm{p_{ti}^o}$ is the direction from projection position $\bm{p_{ti}^{'}}$ to the flow position $\bm{p_{ti}^{o}}$, $\bm{x}=\frac{\partial{e_i}}{\partial{\bm{p_{ti}^{'}}}}$ represents the derivative direction at $\bm{p_{ti}^{'}}$, $\theta$ denotes the angle between $\bm{v}$ and $\bm{x}$, and $\cos(\theta)$ can be computed by
\begin{equation}
\begin{aligned}
\cos(\theta)= \frac{{\bm{v^T}\bm{x}}}{\left|\bm{v}\right|_2\left|\bm{x}\right|_2}.
\end{aligned}
\label{f5}
\end{equation}
As shown in Fig.~\ref{fig:ThetaAndTheta0}, when the projection position $\bm{p_{ti}^{'}}$ lies outside the circle with $\bm{p_{ti}^o}$ as its center and $2\sigma$ as its radius, $\theta_0$ represents the angle between the tangent line $\bm{l}$ and the direction $\bm{v}$. Thus
\begin{equation}
\begin{aligned}
\cos(\theta_0)= \frac{\sqrt{\bm{v^T}\bm{v}-\sigma^2}}{\left|\bm{v}\right|_2}.
\end{aligned}
\label{f08}
\end{equation}

In summary, we tend to activate these correspondences when their projection positions are close to the flow positions or local gradient agrees with global information.

\begin{figure}[t]
  \begin{center}
  \includegraphics[width=0.9\linewidth]{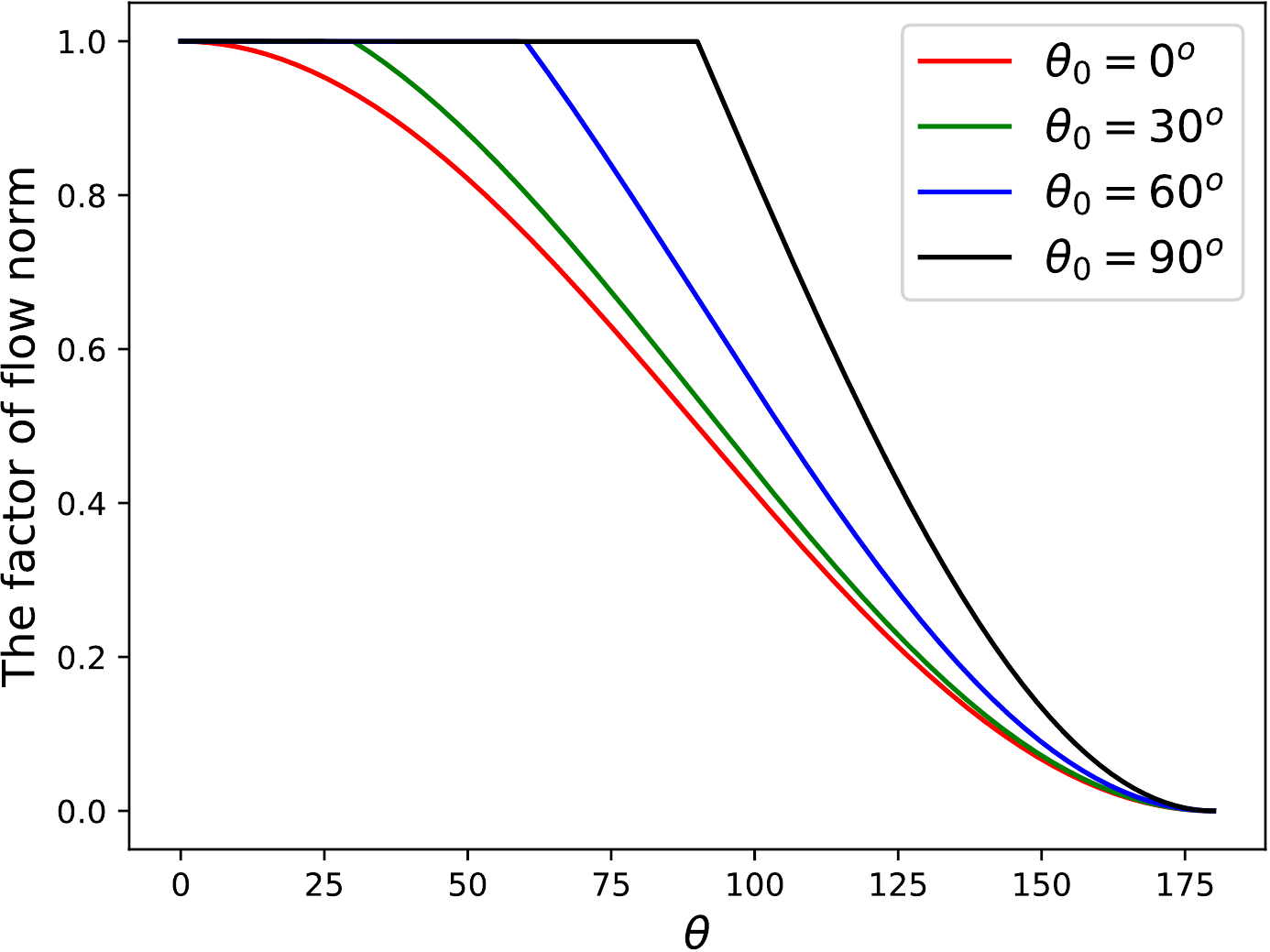}
  \end{center}
\vspace{-0.3cm}
    \caption{Illustration of the change of the flow factor $s$ with the increasing of $\theta$.}
  \label{fig:ThetaChange}
\end{figure}

\begin{figure*}
\begin{center}
%\fbox{\rule{0pt}{2in} \rule{.9\linewidth}{0pt}}
\includegraphics[width=6.8in]{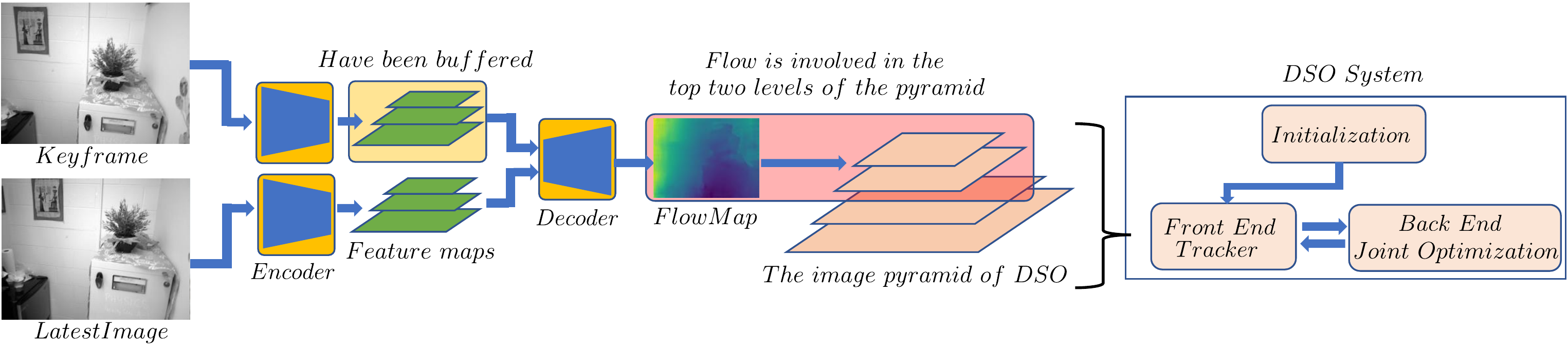}%{0pt}}
\end{center}
   \caption{Overview of the FlowNorm DSO. The predicted flow map is used to suppress those correspondences whose local gradients are highly inconsistent with predicted flow in the top two levels of the image pyramid. }
\label{fig:OverView}
\end{figure*}

With the proposed flow norm, the cost function of direct alignment is formulated as
\begin{equation}
\begin{aligned}
\hat{\bm{\mathcal{X}}}=\mathop{\arg\min}_{\bm{\mathcal{X}}}\sum_{i}^{N}L(\bm{p_{si}},\bm{p_{ti}^{o}},\bm{p_{ti}^{'}}, e_i).
\end{aligned}
\label{f9}
\end{equation}
%where $\bm{p_{si}}$, $\bm{p_{ti}^o}$ and $\bm{p_{ti}^{'}}$ denote the $ith$ item of $\bm{P_s}$,$\bm{P_t^{o}}$ and $\bm{P^{'}_t}$ respectively. 
The new optimal update step of the GN method for the $jth$ iteration is
\begin{equation}
\begin{aligned}
\Delta\widetilde{\bm{\mathcal{X}_j}}=\bm{-(J_j^TSJ_j)^{-1}J_j^TSE_j},
\label{f14}
%\end{split}\]
\end{aligned}
\end{equation}
where $\bm{S}$ is a diagonal matrix, and the $ith$ row and $ith$ column entry is the flow norm factor of the $ith$ residual, which is summarized as
\begin{equation}
s_i = \left\{
             \begin{array}{lr}
             1,  \ \ \ \ \ \ \ \ \               \left|\bm{p_{ti}^o} - \bm{p_{ti}^{'}}\right|_2 \le 2\sigma \\
            	 1, \ \ \ \ \ \ \ \ \ \              \cos{\theta} \le \cos{\theta_0}\ \\ 
             (\frac{\cos{\theta}+1}{\cos{\theta_0}+1}),  \ \ \ \cos\theta_0 < \cos{\theta}.
             \end{array}
\right.
\end{equation}
Fig.~\ref{fig:ThetaChange} illustrates the change of the flow factor $s$ with the increasing of $\theta$. The $\theta_0$ of these four lines is $0\degree$, $30\degree$, $60\degree$ and $90\degree$ respectively.  From Eq.~\eqref{f08}, for the same $\bm{p_{ti}^{'}}$ and $\bm{p_{ti}^{o}}$, a bigger $\theta_0$ corresponds to a bigger flow uncertainty $\sigma$. For a big flow uncertainty, the flow norm will take more account of local information and assign larger weights to it.  In case of the overshoot of the convergence process and the optimized results from being biased by the noise flow, we only involve the flow norm in a tracker when it runs at coarse levels of image pyramid. For example, the image pyramid of DSO has four levels, and we only involve our flow norm in the top two levels.

Although the form of the flow norm is similar to that of the Huber norm, their cores  are very different. The Huber norm utilizes the local information of correspondences and the flow norm depends on the coarse flow. In fact, they are complementary to each other.

\subsection{Shrunken PWC-Net}
\label{Shrinked}
To obtain the optical flow map, we employ a shrunken PWC-Net to predict the optical flow between two images. Approaches that learn to predict optical flow from an image pair have been studied in previous works~\cite{RefWorks:doc:5d7e045de4b0229b68f6b7d6, RefWorks:doc:5d7e0430e4b07e3e85c82f4f, RefWorks:doc:5d7e03f7e4b01d772366adda, RefWorks:doc:5d7e03cce4b064b49fd5626c,RefWorks:doc:5d7f264ce4b0cf4c5f85f9a5}. However, due to the high computation cost, these previous nets cannot be migrated directly to our work. To efficiently obtain the optical flow, we choose the baseline network PWC-Net~\cite{RefWorks:doc:5d7e03cce4b064b49fd5626c} as a reference network, and then shrink its convolutional layers and reduce its input image size. The shrinking process is a tradeoff between prediction accuracy and computing efficiency. As our method works on the coarse levels of the image pyramid, it can robustly utilize the inaccurate optical flow.

Firstly, we change the input size of this network from $[3\times436\times1024]$ to $[1\times240\times320]$. RGB images should be transformed into grey images before feeding them into the shrunken network. Secondly, we remove one coding block and two pooling operations from the encoder, and the output size of the last encoding layer is $[15\times30]$. Finally, we remove one decoding block and reduce the correlation radius from 4 to 3, as the correlation operation of decoding block is computationally expensive. The size of the predicted flow is $[112\times160]$. Our encoding and decoding blocks are identical to the encoding and decoding blocks of PWC-Net. The shrunken network architecture is shown in the supplementary video.

Let $\Theta$ be the set of all the learnable parameters in our shrunken network. $W_{\Theta}^l$  and $W_{GT}^l$ denote the predicted flow field and the corresponding ground truth of the $lth$ pyramid level respectively. We use the same multiscale training loss proposed in FlowNet~\cite{RefWorks:doc:5d7e0430e4b07e3e85c82f4f}: 

\begin{equation}
\begin{aligned}
\mathcal{L}(\Theta)=\sum_{l=l_0}^{L}\alpha_{l}\sum_{x}\left|W_{\Theta}^{l}(x)-W_{GT}^{l}(x)\right|_2 + \gamma|\Theta|_2,
\end{aligned}
\label{f1006}	
\end{equation}
where the second term regularizes the parameters of the model in case of over fitting, the $\alpha_{l}$ and $\gamma$ are the balance weights for different pyramid levels.

The variance $\sigma$ of the predicted flow is computed by averaging the squared $L2$ error of the prediction results in the testing dataset.

%\begin{figure}[t]
%  \begin{center}
%  \includegraphics[width=1.0\linewidth]{OpticalFlow2.pdf}
%  \end{center}
%    \caption{This figure shows two correspondences ($p_0$,$q^j_0$) and ($p_1$,$q^j_1$) at the $jth$ iteration of GN solver.The angles $\theta^j_0$ and $\theta^j_1$ measure whether they are negative correspondences. In this case, $\vec{dq_0^j}$ closes to the real convergence direction $\vec{q_0^j\dot{q_0}}$, which tends to promote the convergence of solver. Conversely, $\vec{dq_1^j}$ is far away from real convergence direction $\vec{q_1^j\dot{q_1}}$, which is a negative correspondence and must be supressing the convergence of solver. $\widehat{q0}, \widehat{q1}$ and error0 and error1 are the prediction postion and prediction erros respectively.}
%  \label{fig:OpticalFlow}
%\end{figure}

%$\overrightarrow{dq_0^j}$
% $\overrightarrow{dq_1^j}$
\subsection{Overview of the FlowNorm DSO}
\label{OverviewSubsection}

To demonstrate the effectiveness and efficiency of our method, we take our flow norm as a plug-in component for the baseline methods, DSO and BA-Net. The FlowNorm DSO is still a real-time system, which can be directly compared with DSO on any dataset. 

As shown in Fig.~\ref{fig:OverView}, the plug-in is composed of three parts: the latest image is firstly fed into the encoder network to construct multi-scale feature maps. Then, the decoder will get the concatenation of the latest feature maps and the keyframe's feature maps and output a predicted flow map. Finally, the predicted flow map will be involved in the BA of the DSO tracker in the top two levels of the image pyramid. Although the pipeline needs to encode two images, all frames are only required to be encoded once by buffering the feature maps of the active keyframes.

%computing initialization from flow
\subsection{Comparison with FlowInit DSO}
\label{Contrast}
To completly prove the effectiveness and efficiency of the flow norm, we also construct a competitive strategy. Given $\bm{p_{si}}\in \bm{P_s}=\{\bm{p_{si}}|i=1\cdots N\}$, $d_i\in D=\{d_i|i=1\cdots N\}$ at $I_s$ and the predicted positions $\bm{p_{ti}^{o}}\in \bm{P_t^{o}}=\{\bm{p_{ti}^{o}}|i=1\cdots N\}$ at $I_t$, we compute the initial transform $\bm{T_0}$ by minimizing the geometric error
\begin{equation}
\\\bm{T_0}=\mathop{\arg\min}_{\bm{T}}\sum_{i}^{N}\left| \pi(\bm{p_i},\bm{T},d_i) - \bm{p_{ti}^{o}} \right|_2,
\label{f14}
\end{equation}
where $\pi(\cdot)$ is the projection function, and it is defined in Sect.~\ref{Revisited}. Then, we take $\bm{T_0}$ as an initialization for the tracker of the DSO. We call the DSO with the initialization computed from the predicted flow map as FlowInit DSO. We find that the performance of this initialization strategy is on par with the FlowNorm DSO strategy for those well predicted flow positions. However, for very poor flow prediction, tracking with the initialization strategy is highly unstable. Because we shrink the size of the prediction network, our predicted flow usually has an overall offset. The overall offset causes that the initialization from the geometric BA also contains the offset. More comparison details are shown in the next Section.

%------------------------------------------------------------------------
\section{Experiments}
To verify the effectiveness of our method, we build FlowNorm versions of DSO and BA-Net. We evaluate our system on a Linux system with an Intel Core i7-7700 CPU of 3.50GHz and an Nvidia Titan Xp GPU.
%meanwhile, we also test time cost of the FlowNorm DSO in Manifold2-C\footnote{https://www.dji.com/cn/manifold-2}, which is a embedded platform.

\subsection{Training}
\label{dataset}
We train the shrunken PWC-Net on the SceneNN~\cite{RefWorks:doc:5c7bd892e4b05f3521397a13} dataset, which consists of 94 Kinect-captured RGB-D image sequences with ground truth poses. We select 44 / 25 image sequences from the SceneNN dataset and take them as training/testing sets respectively. Then, we sample pairs from the training and testing sets and generate the ground truth optical flow by projecting one pixel from one image to another image. During the projection process, we remove the occlusion area by verifying whether the depth of one pixel is consistent with the depth of its projection position. 

Our shrunken PWC-Net is trained with ADAM~\cite{RefWorks:doc:5c7bb56fe4b0ca33023de881} with the initial learning rate 0.0001. The weights in the training loss defined in Eq.~\eqref{f1006} are set to be $\alpha5=0.08$, $\alpha4=0.02$, $\alpha3=0.01$, and $\alpha2=0.005$. The trade-off weight $\gamma$ is set to be 0.0004. Although our network lacks some of the layers of PWC-Net, we still load the parameters of PWC-Net into the corresponding layers of the shrunken version as the initial parameters. The total training process takes one day on a computer with one Titan XP.
 
%
%\begin{figure}[h]
%  \centering
%  \subfigure[Convergence Rate vs Angle Noise]{
%    \label{fig:AngleVsConvergenceRate} %% label for first subfigure
%    \includegraphics[width=0.8\linewidth]{ConvergenceVsAngle.pdf}}
%  \hspace{0.1in}
%  \subfigure[Convergence Rate vs Translation Noise]{
%    \label{fig:TranslationVsConvergenceRate} %% label for second subfigure
%    \includegraphics[width=0.8\linewidth]{ConvergenceVsTranslation.pdf}}
%  \caption{Comparison the convergence rates of enhanced solvers and original solvers for different noise levels, the translation noise level in a) is fixed to $0.1$ and the angle noise level in b) is fixed to $30$.}
%  \label{fig:LinesConvergenceRate} %% label for entire figure
%\end{figure}

\subsection{FlowNorm in DSO}
We compare the FlowNorm DSO with the original DSO on two monocular datasets: TUM-MonoVO dataset~\cite{RefWorks:doc:5d7dfb06e4b054c6b854fc6a} and ICL-NUIM dataset~\cite{RefWorks:doc:5d7dfbd6e4b064b49fd561f2}. The TUM-MonoVO dataset provides 50 photometrically calibrated sequences, comprising different indoors and outdoors environments. The ICL-NUIM dataset contains 8 ray-traced sequences from two indoor environments.
Since the TUM-MonoVO dataset only provides loop-closure ground-truth, we evaluate all sequences using the alignment error, which is defined in the TUM-MonoVO dataset. 
%The code of evaluation method can be found in the website\footnote{https://vision.in.tum.de/data/datasets/mono-dataset} of TUM-MonoVO.
%
%\begin{figure}[h]
%  \centering
%  \subfigure[TUM Mono]{
%    \label{fig:Accuracy1} %% label for first subfigure
%    \includegraphics[width=0.465\linewidth]{Accuracy1.pdf}}
%  \hspace{0.1in}
%  \subfigure[ICL]{
%    \label{fig:Accuracy2} %% label for second subfigure
%    \includegraphics[width=0.445\linewidth]{Accuracy2.pdf}}
%  \caption{The accumulated alignment errors}
%  \label{fig:Accuracy} %% label for entire figure
%\end{figure}

\begin{figure}[h]
  \begin{center}
  \includegraphics[width=0.85\linewidth]{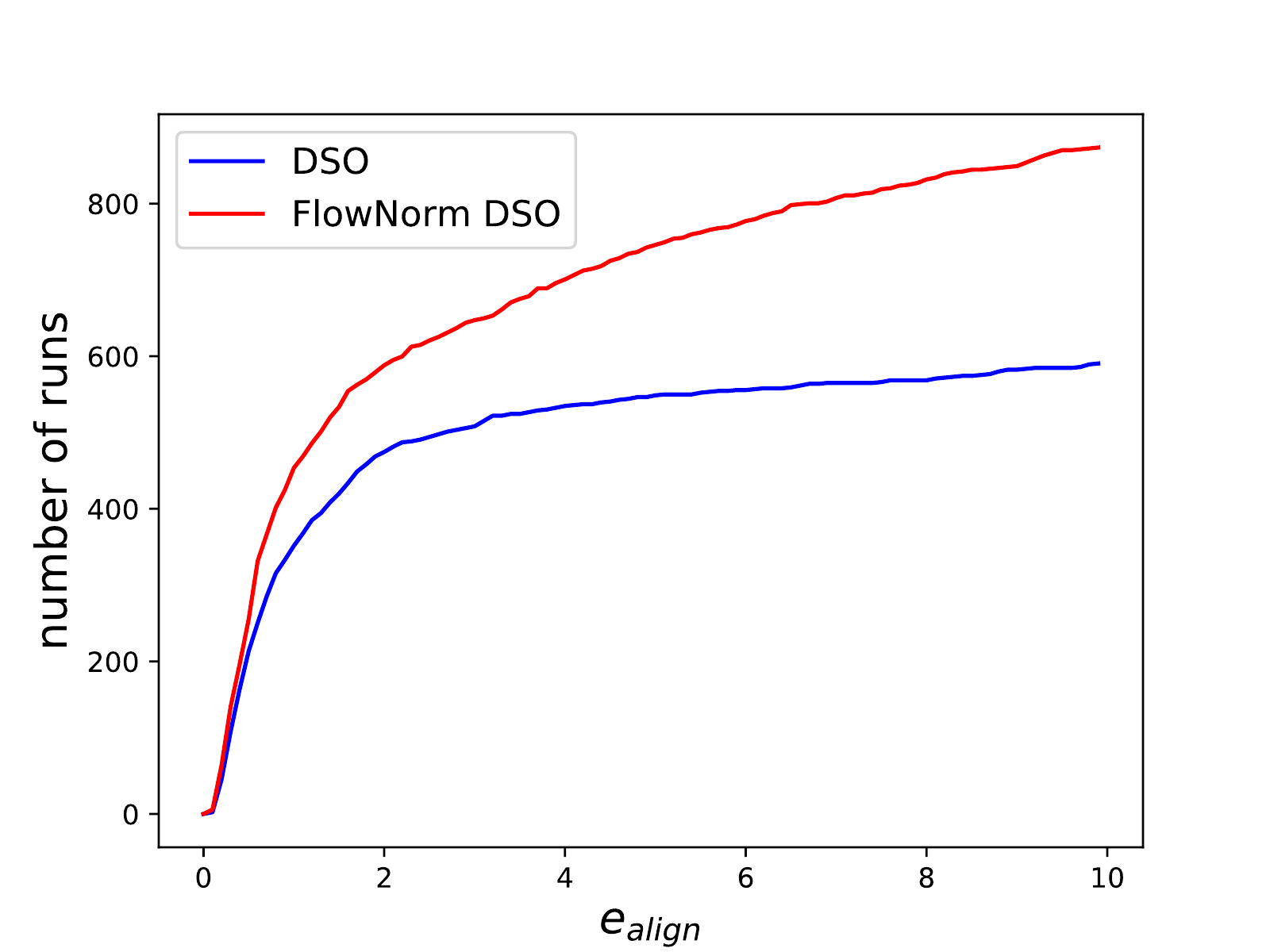}
  \end{center}
\vspace{-0.3cm}
    \caption{The accumulated number of runs whose alignment errors are smaller than $e_{align}$; larger is better. Testing dataset contains all downsampled sequences from TUM-MonoVo and ICL-NUIM datasets.}
  \label{fig:Accuracy}
\end{figure}
\vspace{-0.3cm}
To increase the difficulty of evaluation, we add a {\bf new evaluation metric}. We downsample the image sequences with a skip of 1,2,3,4...13 frames. We track 2 times for every sequence. Which means the total number of runs is 1508.
Apart from the alignment errors, we also measure two numbers: the maximum skip number without a losing tracking and the maximum skip number that can be tracked with an acceptable accuracy for every sequence. We take the alignment error of sequences without downsampling as a reference for whether tracking has acceptable accuracy or not and label it as $error_{0}$. If the alignment error of one downsampling sequence is smaller than three times its corresponding $error_{0}$, we mark the tracking result of the run as an acceptable tracking accuracy. 
\begin{figure}[h]
  \centering
  \subfigure[The maximum skip number with an acceptable tracking accuracy]{
    \label{fig:failure} %% label for first subfigure
    \includegraphics[width=0.85\linewidth]{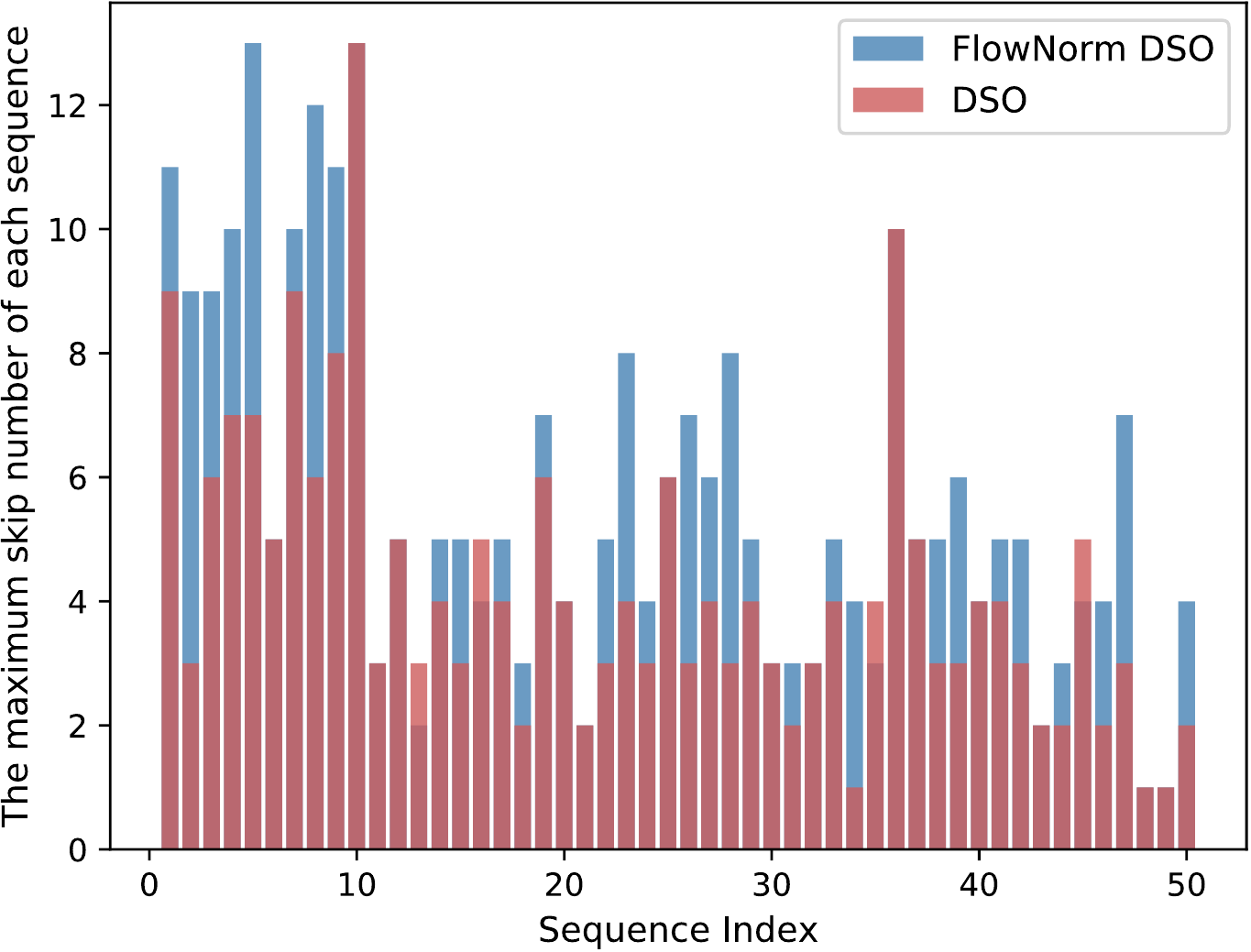}}
  \subfigure[The maximum skip number without a losing tracking]{
    \label{fig:crash} %% label for second subfigure
    \includegraphics[width=0.85\linewidth]{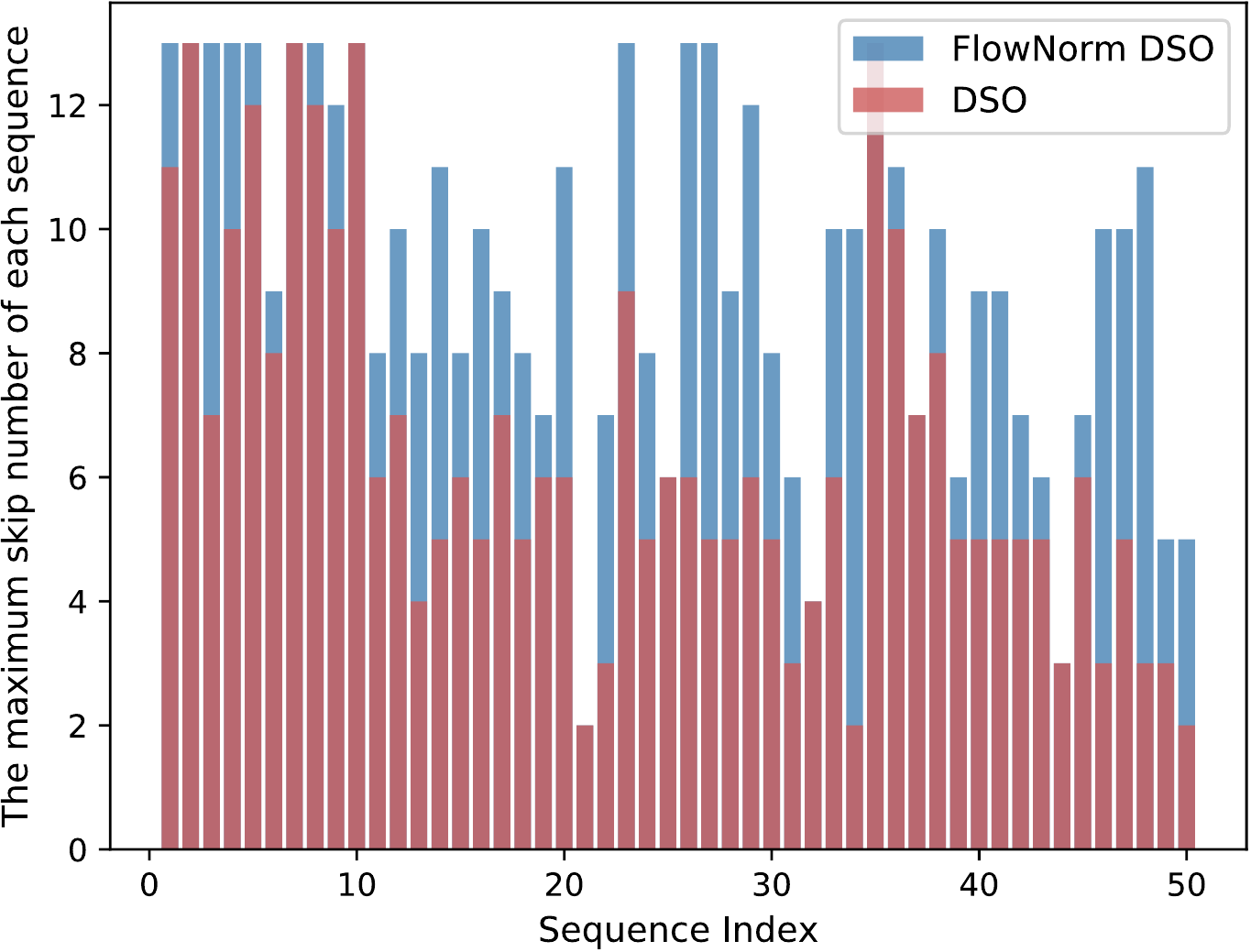}}
  \caption{Comparison of the convergence ability in all 50 sequences of the TUM-MonoVO dataset.}
  \label{fig:bar} %% label for entire figure
\end{figure}

\begin{figure*}
  \begin{center}
  \includegraphics[width=1.0\linewidth]{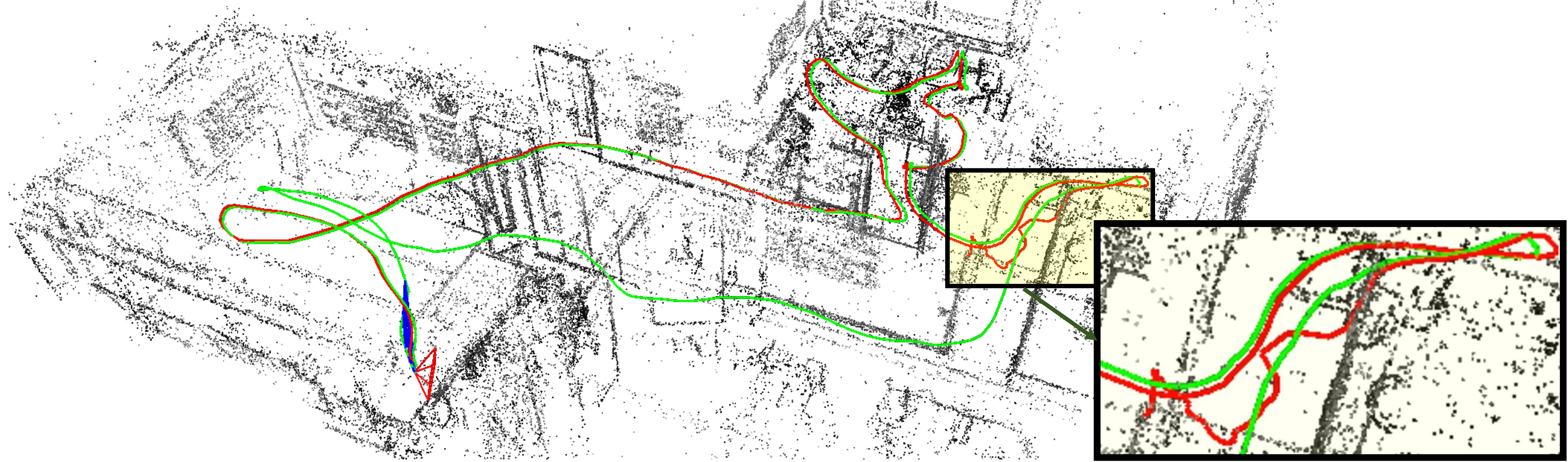}
  \end{center}
    \caption{An example of a losing tracking in the first sequence of the TumMonoVo dataset with a skip of 10 frames. The red and green trajectories are computed from DSO and FlowNorm DSO respectively.}
  \label{fig:Example}
\end{figure*}

Fig.~\ref{fig:Accuracy} illustrates the statistical performance of FlowNorm DSO and DSO. The accuracy of FlowNorm DSO is better than that of the original DSO. The performance of the original DSO is on par with its FlowNorm version when the downsampling rate is low. However, FlowNorm DSO presents more robust performance with the increase of the downsampling rate. Fig.~\ref{fig:bar} shows the maximum skip number with acceptable tracking accuracy and the maximum skip number without losing tracking for all sequences in the TUM Mono dataset. FlowNorm DSO (blue) has consistently better performance than the original version. Note that we only downsample sequences with 1 to 13 steps, the maximum step with 13 means we do not find losing tracking or the tracking results of all runs are acceptable in this sequence.  Fig.~\ref{fig:Example} shows the tracking trajectories of FlowNorm DSO (green) and the original DSO (red) on the first sequence of the TUM Mono dataset with a downsampling rate of 10. DSO loses tracking in the black box area as the camera has large view change there.

\subsection{FlowNorm in BA-Net}
As BA-Net does not public codes, we construct a motion tracking version of it. The constructed BA-Net is trained on our training/validating dataset. Similar to FlowNorm DSO, we also use the predicted flow guide for the convergence of BA-Net. We use the remaining part of the SceneNN dataset to build a challenging image pair dataset. Then, we generate initial poses by adding rotation noise and translation noise to the ground truth poses of these image pairs. The final number of image pairs is 60126. We compare how many pairs are successfully aligned by BA-Net and its FlowNorm version. The results are 48261 and 37218 for FlowNorm version and the original version respectively, which proves our method can further expand the convergence range of the tracker based on minimizing the feature metric residual.

\subsection{FlowNorm Vs FlowInit}
To prove the effectiveness of the flow norm, we construct a competitive strategy, which is described in Sect.~\ref{Contrast}. We compute an initial pose from the predicted flow directly. We find that if we just take the computed initial pose as the initialization of the DSO tracker, the tracker becomes very unstable (usually loses tracking when it gets an inaccurate optical flow). We consider the reason for such losing tracking is that the indirect method seriously depends on the correct matching. However, the depths and correspondences used to compute the initial pose both contain a lot of noise. Next, we insert the computed initial pose to the queue of trying poses in the DSO tracker, and the queue in DSO is used to prevent loss of tracking. We compare the FlowInit DSO and FlowNorm DSO, and the results are shown in Table~\ref{tableFlowVsIntial}. In the table, ``accept. acc." and ``w/o losing." mean the maximum skip number with an acceptable tracking accuracy and the maximum skip number without a losing tracking respectively. ``Ave. align err." denotes the average of the first five alignment errors (downsampling rate from 1 to 5). Due to space limitation, we just show the comparison results for the first three sequences of the TUM-MonoVO dataset.

\begin{table}[h]
\caption{FlowNorm Vs FlowInit}
\vspace{-0.5cm}
\label{tableFlowVsIntial}
\begin{center}
\begin{tabular}{|c|c|c|c|c|}
\hline
Config & Seq. & Ave. align err. & accept. acc.  & w/o losing. \\
\hline\hline
{DSO} &{01} & {0.5760} & 9 & 11 \\
{FlowInit} & {01} & {0.5380} & 9 & {13} \\
{FlowNorm} & {01} & {\textbf{0.5299}} & {\textbf{11}} & {13} \\
\hline
{DSO} & {02} & {1.0094} & 3 & {13} \\
{FlowInit} & {02} & {\textbf{0.3058}} & 8 & {13} \\
{FlowNorm} & {02} & {0.3765} & {\textbf{9}} & {13} \\
\hline
{DSO} & {03} & {0.7237} & 6 & {7} \\
{FlowInit} &{03} & {1.1205} & 7 & {8} \\
{FlowNorm} & {03} & {\textbf{0.6578}} & {\textbf{9}} & {13} \\
\hline
\end{tabular}
\end{center}
\end{table}
\vspace{-0.5cm}

\subsection{Runtime analysis}
In the implementation, we implement the trained model in DSO by PyTorch-C++\footnote{https://pytorch.org/cppdocs/}, and we create a new thread for the flow prediction. The forward of the network is in the GPU and the other models of DSO are implemented in the CPU, which means the prediction of the flow does not have an effect on other models in DSO.  In our computer, the forward process of the shrunken network takes 14 ms per frame. 
%In the Manifold2-C, the forward process cost ..ms per frame.

\section{Conclusion and future work}
In this paper, we have presented a flow norm to enhance the convergence range of direct alignment by utilizing a coarse flow map to constrain those correspondences that are highly inconsistent with the flow map. We employed a shrunken PWC-Net to generate the coarse flow map and built variants of DSO and BA-Net to prove the effectiveness of the flow norm. Meanwhile, we also compared the flow norm with a competitive strategy that gets the initial pose from the predicted flow directly. Our experiments proved the effectiveness and efficiency of the flow norm. In future work, we plan to investigate new network architectures to increase the accuracy of the prediction network and explore more formation of the flow norm.

%, and proved the competitive strategy is unstable for inaccurate flow map
%The testing part is used to construct a challenge dataset for the comparision between the enhanced BA-Net and original BA-Net, the pairs in the dataset have very large view change. 

{\small
\bibliographystyle{unsrt}
\bibliography{egbib}
}

\end{document}